\documentclass{guidelines}

\title{Annotation Guidelines for Corpus {\it Novelties}: Part 2 -- Alias Resolution}
\author{Arthur Amalvy \& Vincent Labatut}
\affiliation{Laboratoire Informatique d'Avignon -- LIA UPR 4128, Avignon, France}
\date{30 September 2024}
\version{1.0.0}

\abstractdef{The \textit{Novelties} corpus is a collection of novels (and parts of novels) annotated for Alias Resolution, among other tasks. This document describes the guidelines applied during the annotation process. It contains the instructions used by the annotators, as well as a number of examples retrieved from the annotated novels, and illustrating how canonical names should be defined, and which names should be considered as referring to the same entity.}

\begin{document}
\maketitle

\newpage
\section{Introduction}
\label{sec:Intro}
This document aims at providing instructions for the annotation of aliases in the \textit{Novelties} corpus. 
The corpus itself 
will be the object of a separate description. 
It was constituted mainly to fulfill two goals: in the short term, train and test NLP methods able to handle \textit{long} texts, and in the longer term, be used to develop \textit{Renard}~\citep{Amalvy2024a}, a pipeline aiming at extracting \textit{character networks} from literary fiction. This pipeline includes several processing steps besides alias resolution, including named entity recognition and coreference resolution. Character networks can be used to tackle a number of tasks, including the assessment of literary theories, the level of historicity of a narrative, detecting roles in stories, classifying novels, identify subplots, segment a storyline, summarize a story, design recommendation systems, align narratives, etc. See the detailed survey of \citet{Labatut2019} for more information regarding character networks. There are seldom annotation guidelines for alias resolution in the literature, so the one presented here are designed from scratch, taking into account this application's context.

\subsection{Notion of \textit{Alias}}
\label{sec:IntroAlias}
In this document, aliases are the different names or expressions used in a novel to refer to the same entity. The goal of alias resolution is to automatically determine whether two expressions refer to the same entity. The task in not well-defined in the literature in general, and is sometimes termed differently. Alternative names include ``character detection'', ``character identification'' or ``character clustering''. While some works are specifically concerned with alias resolution~\citep{Vala2015, Jahan2020}, most solve the task in the pursuit of another one, such as character classification~\citep{Bamman2014} or speaker attribution~\citep{CuestaLazaro2022}. Still, alias resolution is an important step of character network extraction, which is why we choose to include this annotation layer in our corpus.

In the context of \textit{Novelties}, the manual annotation process consists in elaborating a unique name for each entity, called a canonical form, and associate it to each occurrence of the entity. When performing the network extraction, the vertices are named using these canonical forms.

In the rest of the document, we provide a number of examples to illustrate our guidelines. Text extracted from novels is formatted using a sans-serif font, e.g. \textex{the Emperor}.

\subsection{Organization}
\label{sec:IntroOrg}
In the following, we first describe guidelines for the alias annotation process from a very operational perspective (Section~\ref{sec:Process}). We then list the rules used to elaborate the canonical forms, depending on the type of the considered entities (Section~\ref{sec:Canon}). Finally, Section~\ref{sec:Conclu} provides our concluding remarks, and Appendix~\ref{sec:ApdxVersion} gives the history of this document.

\newpage
\section{Annotation Process}
\label{sec:Process}
At this stage, we assume that the considered novel has already been annotated for Named Entity Recognition, using our guidelines~\cite{Amalvy2024f}. It is better if the person that annotated the named entities also performs the alias annotation, as they have more experience with the story and characters, and are less likely to make mistakes.  

The goal of this section is to provide a detailed and very concrete description of the process we use, so that anyone can easily reproduce it. We do not use a specific tool, but rather a spreadsheet editor such as Calc or Excel, and possibly a simple text editor to correct errors in the \texttt{.conll} files. Concretely, the annotation process goes as follows.

\subsection{Preparation of the Resources}
\label{sec:ProcessPrep}
The few operations described in the following can significantly help to perform the whole process, and to reduce the number of mistakes. 

\paragraph{Entity and mention lists}
We use the scripts available on the Novelties GitHub repository\footnote{\url{https://github.com/CompNet/Novelties}} to extract two CSV files:
\begin{enumerate}
    \item Entity list: the list of all unique entity names, with their entity type and frequency;
    \item Mention list: the list of all entity mentions, with their positions in the novel (chapter and line).
\end{enumerate}
The fourth column of the first file is empty, and designed to receive the canonical form of each entity name. The rest of the process consists in filling this column. 

For the sake of convenience, we order this file first by entity name, then by entity type. The former criterion allows grouping relatively similar names, and the latter allows processing the entity types separately. 

\paragraph{Secondary resources}
The next step is to open all necessary secondary resources. These will be used mainly to search additional details, and to perform some verifications. These include:
\begin{itemize}
    \item Some Web content such as a fan Wiki\footnote{e.g. A Wiki of Ice and Fire: \url{https://awoiaf.westeros.org/index.php/Main_Page}} dedicated to the novel, or some annotated version of the text\footnote{e.g. Power Moby Dick: \url{http://www.powermobydick.com/}}. This is useful to determine the canonical name of certain entities.
    \item The epub version of the novel, which can be viewed in a software such as Calibre\footnote{\url{https://calibre-ebook.com/}}. This allows efficiently searching through the book to make various verifications.
    \item The list of entity mentions extracted at the previous step. It can be opened using a spreadsheet editor, but does not require any sorting, unlike the list of entity names. It can be convenient to filter the rows by entity type, though.
\end{itemize}

\subsection{Processing of an Entity Name}
\label{sec:ProcessProc}
For each entity name in the list, we perform the following operations.

\paragraph{Fill the canonical form} 
In the CSV file (entity list), we add the canonical form associated to the current entity name. Very often, this amounts to pasting the same name. For instance, if one is processing \textex{Harry Potter}, then one will use the exact same string as the canonical form. 

\paragraph{Check suspect names}
If the current name looks non-canonical, or if it could be an alias in the strict sense (i.e. a name completely different of the character's real name), it is important to check in the novel and in the secondary sources. For instance, if one is processing \textex{Voldemor}, then one needs to retrieve the canonical form \textex{Tom Marvolo Riddle} from a secondary source. The rules to determine the canonical form are described in Section~\ref{sec:Canon}.

\paragraph{Search for other occurrences} 
Once the canonical form is set, we look for parts of the entity name in the rest of the CSV file, in order to find other names referring to the same entity. For instance, if the current name can be split into a first name and a last name (ex. \textex{Harry Potter}), one can look for both separately (\textex{Harry} and \textex{Potter}). If the other entity names already have a canonical form, it is important to verify that they are exactly the same as the current one. If they do not, then one can associate the current canonical form to these other entity names. 

\paragraph{Correct annotation errors}
Sometimes, the current entity name looks incorrect: it is visibly not a character name (\textex{autobus}), or it contains non-alphabetic characters (\textex{Potter!!!}), or it seems incomplete (\textex{Potte}), etc. In this case, there is probably an error in one of the \texttt{.conll} files containing the NER annotation. To correct this mistake, we use the mention list (i.e. the other CSV file) to precisely locate the error, and directly edit the \texttt{.conll} file. We also correct the problem in the entity list (but not in the mention list), for future verification.

\subsection{Overall Verification}
\label{sec:ProcessVerif}
Once all the names in the entity list have been processed as described before, we get what we call a v1 of the alias annotations. It is now necessary to perform certain verifications in this list. 

\paragraph{Unicity of the Canonical Forms}
First, we verify that the same canonical form is not written in several different ways (e.g. minor character case differences). In the now annotated entity list, generate the list of unique names. When using LibreOffice Calc:
\begin{enumerate}
    \item Select the column containing canonical forms; 
    \item Go to menu \textit{Data} > \textit{More Filters} > \textit{Standard Filter}
    \item In the dialog box, \textit{Filter Criteria}, set \textit{Field Name} to \texttt{none};
    \item In \textit{Options}, check \textit{No duplications} and \textit{Copy results to}, then select the target column;
    \item Uncheck \textit{Keep filter criteria}
    and \textit{Case sensitive}
\end{enumerate}
If a canonical form presents any variability, we correct this.

\paragraph{Regeneration of the Lists}
As in the first step (Section~\ref{sec:ProcessPrep}), we use the scripts to generate the entity list and mention list. While doing so, we must make sure to keep the v1 file intact, as we need both versions of the entity list. In the following, v2 refers to the updated entity list.

\paragraph{Verification of the Entity Annotations}
This operation is particularly important, especially if some corrections have been conducted in the \texttt{.conll} files during the annotation process. We copy the content of the v1 file and paste it in the v2 file, near the existing content. In an empty column, we now define a formula that checks whether the v1 canonical form is equal to the v2 canonical form. We paste the formula over all rows. Finally, we check whether all the values in this new column are \texttt{TRUE}. Any \texttt{FALSE} value reveals a difference between v1 and v2, generally corresponding to a modification incorrectly performed during the second step (Section~\ref{sec:ProcessProc}). It is necessary to properly apply them, by performing the appropriate correction in the \texttt{.conll} files. 

\medskip
After these corrections, we must do the verification process again: regenerate the entity list, and compare it to v1. Only when both lists are exactly similar can we conclude the whole process.

\subsection{Metadata and Finalization}
\label{sec:ProcessFinal}
Once we reach a stable annotation file, we create a new, fifth, column, entitled \texttt{metadata}, in order to store the following information, similarly to what we do in the NER annotation files~\cite{Amalvy2024f}:
\begin{itemize}
    \item Title of the novel (field \texttt{Title});
    \item Name of the annotators (field \texttt{Annotator});
    \item Version of the guidelines (field \texttt{Guidelines});
    \item Date of last update (field \texttt{Updated}).
\end{itemize}

Finally, we rename the file as \texttt{alias\_resolution.csv} and place it in the same folder as the NER annotation files.

\newpage
\section{Canonical Forms}
\label{sec:Canon}
We follow a certain number of rules to define the canonical form of an entity's name. The general idea is that it should be complete enough to clearly identify the entity unambiguously. Or at least, as much as possible depending on the novel and entity. These rules are slightly different from one entity type to the other. It is important to stress that the canonical form does not necessarily appear in the original text. For example, it can come from an external source.

\subsection{Character Entities (\texttt{CHR})}
\label{sec:CanonChr}

\paragraph{General Rule}
The general rule for character entities is to form the canonical name based only on the first and family names. See for instance \textex{Constance Bonacieux}, in Alexandre Dumas's \textit{The Three Musketeers}:

\smallskip\noindent\begin{tabularx}{\textwidth}{X X}
    \hline
    \rowcolor{lightgray!50} \textbf{Form in the text} & \textbf{Canonical form} \\
    \hline
    \textex{Constance} & Constance Bonacieux \\
    \textex{Constance Bonacieux} & \\
    \textex{Madame Bonacieux} & \\
    \textex{Mme. Bonacieux} & \\
    \hline
\end{tabularx}

\smallskip\noindent We ignore honorifics (here \textex{Madame} and \textex{Mme.}), unless the first name or the last name is unknown. In this case, we include the main honorific title, for instance:

\smallskip\noindent\begin{tabularx}{\textwidth}{X X}
    \hline
    \rowcolor{lightgray!50} \textbf{Form in the text} & \textbf{Canonical form} \\
    \hline
    \textex{Comte de Wardes} & Comte de Wardes \\
    \textex{De Wardes} & \\
    \textex{M. de Wardes} & \\
    \textex{Monsieur de Wardes} & \\
    \textex{Monsieur le Comte de Wardes} & \\
    \hline
    \textex{Coquenard} & Monsieur Coquenard \\
    \textex{M. Coquenard} & \\
    \textex{Monsieur Coquenard} & \\
    \hline
\end{tabularx}

\paragraph{Tripartite Names}
Certain cultures use names composed of three parts, in which case our canonical form includes all three of them. This is the case, in particular, of Latin (example from \textit{The Three Musketeers}) and Russian (example from Dostoevsky's \textit{The Double}) names:

\smallskip\noindent\begin{tabularx}{\textwidth}{X X}
    \hline
    \rowcolor{lightgray!50} \textbf{Form in the text} & \textbf{Canonical form} \\
    \hline
    \textex{Caesar} & Gaius Julius Caesar \\
    \hline
    \textex{Yakov Petrovitch Golyadkin} & Yakov Petrovitch Golyadkin \\
    \textex{Yakov Petrovitch} & \\
    \textex{brother Yakov} & \\
    \textex{Mr. Golyadkin} & \\
    \hline
\end{tabularx}

Other types of novels are likely to exhibits different types of names, e.g. completely made-up systems in Fantasy novels, or context-dependent names in the Chinese culture\footnote{\url{https://en.wikipedia.org/wiki/Chinese_name\#Alternative_names}}. The general principle here is to adapt the nature of the canonical names to the considered novel.

\paragraph{Aliases}
If a character appears under completely different aliases, we try to use their actual name, e.g. for \textex{D'Artagnan}'s antagonist \textex{Milady} in \textit{The Three Musketeers}:

\smallskip\noindent\begin{tabularx}{\textwidth}{X X}
    \hline
    \rowcolor{lightgray!50} \textbf{Form in the text} & \textbf{Canonical form} \\
    \hline
    \textex{Anne de Breuil} & Anne de Breuil \\
    \textex{Charlotte Backson} & \\
    \textex{Comtesse de la Fère} & \\
    \textex{Comtesse de Winter} & \\
    \textex{Lady Clarik} & \\
    \textex{Milady} & \\
    \textex{Milady Clarik} & \\
    \textex{Milady de Winte}r & \\
    \hline
\end{tabularx}

\paragraph{Nicknames}
When the character possesses a nickname that is important to its identification, it is included in the canonical form as a complement. For instance, in \textit{The Three Musketeers}, the main character is only known as \textex{D'Artagnan}, and one of the musketeers is simply Athos:

\smallskip\noindent\begin{tabularx}{\textwidth}{X X}
    \hline
    \rowcolor{lightgray!50} \textbf{Form in the text} & \textbf{Canonical form} \\
    \hline
    \textex{D'Artagnan} & Charles de Batz de Castelmore, dit d'Artagnan \\
    \textex{Lord d'Artagnan} & \\
    \textex{M. d'Artagnan} & \\
    \textex{Monsieur d'Artagnan} & \\
    \hline
    \textex{Athos} & Olivier de La Fère, dit Athos \\
    \textex{Comte de la Fère} & \\
    \textex{Monsieur Athos} & \\
    \hline
\end{tabularx}

\smallskip\noindent As mentioned before, the canonical form does not necessarily appear in the original text: it may come from a secondary source, as is the case here.

\paragraph{Historical Characters}
Certain authors like to mention historical characters, to provide some context to their story. We retrieve the full name from secondary sources, even when they are not used in the novel. For instance, in \textit{The Three Musketeers}:

\smallskip\noindent\begin{tabularx}{\textwidth}{X X}
    \hline
    \rowcolor{lightgray!50} \textbf{Form in the text} & \textbf{Canonical form} \\
    \hline
    \textex{Brutus} & Marcus Junius Brutus \\
    \hline
    \textex{Cervantes} & Miguel de Cervantes \\
    \hline
    \textex{Robespierre} & Maximilien de Robespierre \\
    \hline
\end{tabularx}

\paragraph{Nobility Ranks}
We proceed similarly when the character possesses a nobility rank mainly used to identify them. Consider for example one of the main antagonists, \textex{Cardinal Richelieu}:

\smallskip\noindent\begin{tabularx}{\textwidth}{X X}
    \hline
    \rowcolor{lightgray!50} \textbf{Form in the text} & \textbf{Canonical form} \\
    \hline
    \textex{Cardinal de Richelieu} & Armand Jean du Plessis, duc de Richelieu \\
    \textex{Cardinal Richelieu} &  \\
    \textex{M. de Richelieu} &  \\
    \textex{Monseigneur the Cardinal} &  \\
    \textex{Monseigneur the Cardinal Richelieu} &  \\
    \hline
\end{tabularx}

\paragraph{Kings and Queens}
When dealing with a king or queen, we include the kingdom in the canonical form, in order to avoid any confusion with other monarchs with the same name. Take \textex{Francis the First}, which is mentioned in \textit{The Three Musketeers}: 

\smallskip\noindent\begin{tabularx}{\textwidth}{X X}
    \hline
    \rowcolor{lightgray!50} \textbf{Form in the text} & \textbf{Canonical form} \\
    \hline
    \textex{Francis I} & Francis I of France \\
    \textex{Francis the First} & \\
    \hline
\end{tabularx}

\smallskip\noindent The novel refers to the king of France, and not to the Emperor of Austria bearing the same name.

\paragraph{God \& Satan}
The Abrahamic god appears under a number of names, which we all associate to the canonic form \textex{God}. In \textit{The Three Musketeers}, for instance:

\smallskip\noindent\begin{tabularx}{\textwidth}{X X}
    \hline
    \rowcolor{lightgray!50} \textbf{Form in the text} & \textbf{Canonical form} \\
    \hline
    \textex{Dieu} & God \\
    \textex{Father} & \\
    \textex{Gad} & \\
    \textex{God} & \\
    \textex{Holy Father} & \\
    \textex{Holy Ghost} & \\
    \textex{Son} & \\
    \hline
\end{tabularx}

\smallskip\noindent Note that we also annotate the Catholic trinity as \textex{God}, for the sake of simplicity.

We proceed similarly with Satan and all his names. For instance, in Herman Melville's \textit{Moby Dick}:

\smallskip\noindent\begin{tabularx}{\textwidth}{X X}
    \hline
    \rowcolor{lightgray!50} \textbf{Form in the text} & \textbf{Canonical form} \\
    \hline
    \textex{Beelzebub} & Satan \\
    \textex{Devil} & \\
    \textex{Evil One} & \\
    \textex{Lucifer} & \\
    \textex{Prince of the Powers of the Air} & \\
    \textex{Satan} & \\
    \hline
\end{tabularx}

\subsection{Location Entities (\texttt{LOC})}
\label{sec:CanonLoc}

\paragraph{General Rule}
For locations, the general rule is straightforward, and consists in just using the proper noun as the canonical form:

\smallskip\noindent\begin{tabularx}{\textwidth}{X X}
    \hline
    \rowcolor{lightgray!50} \textbf{Form in the text} & \textbf{Canonical form} \\
    \hline
    \textex{America} & America \\
    \hline
    \textex{Amiens} & Amiens \\
    \hline
    \textex{Angers} & Angers \\
    \hline
    \textex{Angoutin} & Angoutin \\
    \hline
    \textex{Anjou} & Anjou \\
    \hline
\end{tabularx}

\paragraph{Noun Modifiers}
However, sometimes the proper noun is ambiguous when considered separately, and one needs to add a modifier. For instance, in \textit{The Three Musketeers}:

\smallskip\noindent\begin{tabularx}{\textwidth}{X X}
    \hline
    \rowcolor{lightgray!50} \textbf{Form in the text} & \textbf{Canonical form} \\
    \hline
    \textex{Abbey St. Germain} & Abbey St. Germain \\
    \hline
    \textex{St. Germain} & Faubourg St. Germain \\
    \hline
\end{tabularx}

\smallskip\noindent Here, the first example refers to a monastery, whereas the second one is a suburb. Even when there is no ambiguity, it is better to keep a modifier if it helps to understand what the name refers to, e.g. \textex{Mediterranean sea}, \textex{Mount Kilimanjaro}. This makes it easier to identify what the entity is, even without needing to check its type (e.g. \texttt{LOC}).

Expressions using \textex{road} are a bit specific. See these examples from \textit{The Three Musketeers}:

\smallskip\noindent\begin{tabularx}{\textwidth}{X X}
    \hline
    \rowcolor{lightgray!50} \textbf{Form in the text} & \textbf{Canonical form} \\
    \hline
    \textex{road to Picardy} & Picardy \\
    \hline
    \textex{road of Chaillot} & Road of Chaillot \\
    \hline
\end{tabularx}

\smallskip\noindent The context for the first example above is: ``\textex{While his Eminence was seeking for me in Paris, I would take, without sound of drum or trumpet, the road to Picardy, and would go and make some inquiries concerning my three companions.}'' We understand that \textex{road to Picardy} refers to a transportation means rather than a place, and what is important here is the destination, i.e. \textex{Picardy}. The context for the second example is: ``\textex{Between six and seven o’clock the road of Chaillot is quite deserted; you might as well go and ride in the forest of Bondy.}'' Here, the author specifically refers to the \textex{road of Chaillot} as a location, where some event takes place.

\paragraph{Part of Locations}
Sometimes, the proper noun of a location is completed in order to refer to a part of this location. In this case, we consider the relevance of the smaller location in the story: is it important to make the distinction with the larger location? If the answer is yes, then we keep the precision, otherwise, we use the same canonical form as for the larger entity. For instance, in \textit{Moby Dick}:

\smallskip\noindent\begin{tabularx}{\textwidth}{X X}
    \hline
    \rowcolor{lightgray!50} \textbf{Form in the text} & \textbf{Canonical form} \\
    \hline
    \textex{Banks of Newfoundland} & Newfoundland \\
    \hline
    \textex{coast of Greenland} & Greenland \\
    \hline
\end{tabularx}

\subsection{Organization Entities (\texttt{ORG})}
\label{sec:CanonOrg}

\paragraph{Political Entities}
In order to clearly distinguish organizations from locations, we explicitly indicate the nature of the organization in the canonical form, even if it is not specified in the novel. In \textit{The Three Musketeers}:

\smallskip\noindent\begin{tabularx}{\textwidth}{X X}
    \hline
    \rowcolor{lightgray!50} \textbf{Form in the text} & \textbf{Canonical form} \\
    \hline
    \textex{Austria} & Archduchy of Austria \\
    \hline
    \textex{Denmark} & Kingdom of Denmark \\
    \hline
    \textex{France} & Kingdom of France \\
    \textex{kingdom of France} & \\
    \hline
    \textex{Paris} & City of Paris \\
    \hline
\end{tabularx}

\paragraph{Commercial Entities}
It is sometimes not clear that an entity is a commercial structure based on its name alone, so we generally include a precision in its canonical form. In \textit{The Three Musketeers}: 

\smallskip\noindent\begin{tabularx}{\textwidth}{X X}
    \hline
    \rowcolor{lightgray!50} \textbf{Form in the text} & \textbf{Canonical form} \\
    \hline
    \textex{Golden Lily} & Inn of the Golden Lily \\
    \hline
    \textex{Jolly Miller} & Hostel of the Jolly Miller \\
    \hline
    \textex{Post} & Tavern of the Post \\
    \hline
\end{tabularx}

\subsection{Group Entities (\texttt{GRP})}
\label{sec:CanonGrp}

\paragraph{Families}
In order to be very clear that a family name does not refer to a specific person but to the whole group, we explicitly add \textex{House} to the canonical form. For instance, in \textit{The Three Musketeers}:

\smallskip\noindent\begin{tabularx}{\textwidth}{X X}
    \hline
    \rowcolor{lightgray!50} \textbf{Form in the text} & \textbf{Canonical form} \\
    \hline
    \textex{Condés} & House Condé \\
    \hline
    \textex{Montmorency} & House of Montmorency \\
    \hline
\end{tabularx}

\smallskip\noindent We use the word \textex{House} because it is explicitly used in this novel. But in a more general case, one could use \textex{Family} instead: the most important point is to be consistent over the whole novel.

\paragraph{Demonyms \& Ethnonyms}
When dealing with group names that reflect a geographic or ethnical origin, we use the plural form. If no such form exists in English, then we add \textex{people} to the adjective. See these examples from \textit{The Three Musketeers}: 

\smallskip\noindent\begin{tabularx}{\textwidth}{X X}
    \hline
    \rowcolor{lightgray!50} \textbf{Form in the text} & \textbf{Canonical form} \\
    \hline
    \textex{Andalusian} & Andalusians \\
    \hline
    \textex{Arabian} & Arabs \\
    \hline
    \textex{Assyrians} & Assyrians \\
    \hline
    \textex{Béarnais} & Béarnese people \\
    \textex{Béarnese} & \\
    \hline
    \textex{Berrichan} & Berrichons \\
    \textex{Berrichon} & \\
    \hline
    \textex{English} & Englishmen \\
    \textex{Englishman} & \\
    \textex{Englishmen} & \\
    \textex{Englishwoman} & \\
    \hline
\end{tabularx}

\smallskip\noindent If there are several different forms distinguishing gender, we adopt the majority rule, and use the one with the most occurrences as the canonical form. In the above example, \textex{Englishman} (36 occurrences) and \textex{Englishmen} (10) are more frequent than \textex{Englishwoman} (5), which is why we use \textex{Englishmen} as the canonical form.

\subsection{Miscellaneous Entities (\texttt{MSC})}
\label{sec:CanonMsc}

\paragraph{Food}
When the entity refers to food, we generally specify it in the canonical form, especially if it can be confused with another type of entity (such as a location). In \textit{The Three Musketeers}:

\smallskip\noindent\begin{tabularx}{\textwidth}{X X}
    \hline
    \rowcolor{lightgray!50} \textbf{Form in the text} & \textbf{Canonical form} \\
    \hline
    \textex{Bordeaux} & Bordeaux wine \\
    \textex{Burgundy} & Burgundy wine \\
    \textex{chambertin} & Chambertin wine \\
    \textex{champagne} & Champagne wine \\
    \hline
\end{tabularx}

\smallskip\noindent \textex{Bordeaux} is a city, and \textex{Burgundy} and \textex{Champagne} are regions, in France.

\paragraph{Events and Periods}
The principle is the same when dealing with historical events or periods: we want to avoid any confusion. In \textit{The Three Musketeers}:

\smallskip\noindent\begin{tabularx}{\textwidth}{X X}
    \hline
    \rowcolor{lightgray!50} \textbf{Form in the text} & \textbf{Canonical form} \\
    \hline
    \textex{Chalais} & Henri de Talleyrand, marquis de Chalais \\
    \hline
    \textex{Chalais} & Chalais's conspiracy \\
    \hline
\end{tabularx}

\smallskip\noindent Here, the Conspiracy of Chalais was a plot against Richelieu that took place before the story told in \textit{The Three Musketeers}. But Chalais is also the name of the most central person in this plot.

Regarding periods, we also complete the name that appears in the novel, if necessary. Here are some example from Glen Cook's \textit{The Black Company}:

\smallskip\noindent\begin{tabularx}{\textwidth}{X X}
    \hline
    \rowcolor{lightgray!50} \textbf{Form in the text} & \textbf{Canonical form} \\
    \hline
    \textex{ancient kingdom} & Ancient Kingdom period \\
    \hline
    \textex{Cho'n Delor} & Cho’n Delor era \\
    \hline
\end{tabularx}

\paragraph{Works}
For intellectual and artistic works, we build the canonical form based on the author's name and the work's title. In \textit{The Three Musketeers}:

\smallskip\noindent\begin{tabularx}{\textwidth}{X X}
    \hline
    \rowcolor{lightgray!50} \textbf{Form in the text} & \textbf{Canonical form} \\
    \hline
    \textex{Augustinus} & C. Jansenius's \textit{Augustinus} \\
    \hline
    \textex{Iliad} & Homer's \textit{Iliad} \\
    \hline
    \textex{L'Avare} & Molière's \textit{L'avare} \\
    \hline
\end{tabularx}

\smallskip\noindent When the text refers to someone's corpus of work more globally, we use the name of the author and mention its whole work:

\smallskip\noindent\begin{tabularx}{\textwidth}{X X}
    \hline
    \rowcolor{lightgray!50} \textbf{Form in the text} & \textbf{Canonical form} \\
    \textex{St. Augustine} & St. Augustine's work \\
    \hline
    \textex{St. Bartholomew} & Bartholomew's work \\
    \hline
    \textex{St. Chrysostom} & St. Chrysostom's work \\
    \hline
\end{tabularx}

\smallskip\noindent If the author is unknown, we simply indicate the nature of the work, like in these examples from Aldous Huxley's \textit{Brave New World}:

\smallskip\noindent\begin{tabularx}{\textwidth}{X X}
    \hline
    \rowcolor{lightgray!50} \textbf{Form in the text} & \textbf{Canonical form} \\
    \textex{Hug me till you drug me, honey} & \textit{Hug me till you drug me, honey} song \\
    \textex{Three Weeks in a Helicopter} & \textit{Three Weeks in a Helicopter} movie \\
    \hline
\end{tabularx}

\paragraph{Scriptures}
We use \textex{Bible} as the canonical form for all parts of the Bible. In \textit{The Three Musketeers}, for instance:

\smallskip\noindent\begin{tabularx}{\textwidth}{X X}
    \hline
    \rowcolor{lightgray!50} \textbf{Form in the text} & \textbf{Canonical form} \\
    \hline
    \textex{Bible} & Bible \\
    \textex{Scripture} & \\
    \textex{Scriptures} & \\
    \textex{Judith} & \\
    \hline
\end{tabularx}

\smallskip\noindent Here, \textex{Judith} refers to the \textit{Book of Judith} in the \textit{Old Testament}.

\paragraph{Languages}
We explicitly state the nature of a language in its canonical form, to avoid any confusion with demonyms. In \textit{The Three Musketeers}:

\smallskip\noindent\begin{tabularx}{\textwidth}{X X}
    \hline
    \rowcolor{lightgray!50} \textbf{Form in the text} & \textbf{Canonical form} \\
    \hline
    \textex{English} & English language \\
    \hline
    \textex{French} & French language \\
    \hline
    \textex{German} & German language \\
    \hline
\end{tabularx}

\newpage
\section{Concluding Remarks}
\label{sec:Conclu}
The annotation of aliases is much more time-consuming than one would expect \textit{a priori}. Indeed, it requires performing many verifications in order to make sure that we get a proper bijection between the sets of entities and surface forms. This is particularly true for novels taking place in historical settings (e.g. \textit{The Three Musketeers}, \textit{Eugénie Grandet}, \textit{The Red and the Black}), as one must recover the full names of certain entities, that are not always mentioned in the book. Moreover, these novels mix fictional and real entities, that must also be checked against some reliable reference (e.g. Wikipedia, or a specialized Wiki). Novels such as \textit{Moby Dick}, with many cultural references, are also hard: their annotation is significantly eased through the use of a commented version of the novel. Generative models can also be of some help, especially when dealing with historical references in classic novels, although at the time of writing, their answers must be carefully verified. We typically provide the model with an excerpt of the novel and ask it about the specific entity.

Another time-consuming aspect of this annotation tasks, is that it reveals a number of issues among the NER annotations. This mainly concerns the boundaries of the annotations (too many words, too few words, \texttt{B-xxx} instead of \texttt{I-xxx} or the opposite); their type (missing type, incorrect type); and inconsistencies in the NER annotation process throughout the novel (e.g. some definite description not annotated systematically). In any case, detecting such issues implies correcting the NER annotation, and performing a new verification of the aliases. 

Determining whether two surface forms refer to the same entity can be quite easy when one form is just a variant of the other (e.g. \textex{D'Artagnan} vs. \textex{M. D'Artagnan}). But it can also be very tricky, when one entity has completely different names (cf. \textex{Milady} in \textit{The Three Musketeers}, Section~\ref{sec:CanonChr}). The latter case requires a relatively good understanding of the story. For this reason, it is much more convenient, easy, and efficient, that 1) the same person defines the NER and alias annotations; and 2) the alias annotation is conducted right after the NER annotation, when the story is still fresh in the mind of the annotator.

\appendix
\newpage
\section{Version History}
\label{sec:ApdxVersion}
We use three-part version numbers of the form major--minor--patch for both these guidelines and the \textit{Novelties} corpus. See the named entity annotation guidelines for more details~\cite{Amalvy2024f}.

\begin{center}
    \begin{tabularx}{\linewidth}{l l X p{1.3cm}}
        \hline
        \textbf{Version} & \textbf{Date} & \textbf{Changes} & \textbf{Corpus} \\
        \hline
        \texttt{1.0.0} & 30/09/2024 & First version, based on the annotation of a few novels. \\
        \hline
    \end{tabularx}
\end{center}

\paragraph{Version \texttt{1.0.0}}
This is the first version of our guidelines for alias resolution. It is based on the annotation of a few books including fantasy (\textit{The Blade Itself}, \textit{The Black Company}), Science-Fiction (\textit{1984}, \textit{Brave New World}) and classic (\textit{Moby Dick}, \textit{The Three Musketeers}, \textit{Eugénie Grandet}) novels.


\newpage
\phantomsection\addcontentsline{toc}{section}{References}


\footnotesize
\setlength{\bibsep}{3pt plus 1.5ex}
\bibliography{novelties_biblio.bib}

\end{document}